\def\year{2019}\relax
\documentclass[letterpaper]{article} 
\usepackage{aaai19}  
\usepackage{times}  
\usepackage{helvet}  
\usepackage{courier}  
\usepackage{url}  
\usepackage{graphicx}  
\usepackage{latexsym}

\usepackage[utf8]{inputenc} 
\usepackage[T1]{fontenc}    
\usepackage{wrapfig}
\usepackage{cleveref}       
\usepackage{url}            
\usepackage{booktabs}       
\usepackage{amsfonts}       
\usepackage{nicefrac}       
\usepackage{microtype}      
\usepackage{multirow}
\usepackage{url}
\usepackage{graphicx}
\usepackage{color}
\usepackage{tabularx}
\usepackage{amsmath}
\usepackage{caption,subcaption}
\usepackage{wrapfig}

\newcommand*{\affaddr}[1]{#1}
\newcommand*{\affmark}[1][*]{\textsuperscript{#1}}
\newcommand*{\email}[1]{\texttt{#1}}

\begin{document}
%
\title{Latent Multi-task Architecture Learning}
\author{Sebastian Ruder\affmark[1]\affmark[2], Joachim Bingel\affmark[3], Isabelle Augenstein\affmark[3], Anders S{\o}gaard\affmark[3]\\
\affaddr{\affmark[1]Insight Research Centre, National University of Ireland, Galway}\\
\affaddr{\affmark[2]Aylien Ltd., Dublin, Ireland}\\
\affaddr{\affmark[3]Department of Computer Science, University of Copenhagen, Denmark}\\
\email{sebastian@ruder.io},  \email{\{bingel|augenstein|soegaard\}@di.ku.dk}
}
\maketitle
\begin{abstract}
Multi-task learning (MTL) allows deep neural networks to learn from related tasks by sharing parameters with other networks. In practice, however, MTL involves searching an enormous space of possible parameter sharing architectures to find (a) the layers or subspaces that benefit from sharing, (b) the appropriate amount of sharing, and (c) the appropriate relative weights of the different task losses. Recent work has addressed each of the above problems in isolation. In this work we present an approach that learns a \emph{latent} multi-task architecture that jointly addresses (a)--(c). We present experiments on synthetic data and data from OntoNotes~5.0, including four different tasks and seven different domains. Our extension consistently outperforms previous approaches to learning latent architectures for multi-task problems and achieves up to 15\%~average error reductions over common approaches to MTL.
\end{abstract}

\section{Introduction}
Multi-task learning (MTL) in deep neural networks is typically a result of parameter sharing between two networks (of usually the same dimensions) \cite{Caruana:93}. If you have two three-layered, recurrent neural networks, both with an embedding inner layer and  each recurrent layer feeding the task-specific classifier function through a feed-forward neural network, we have 19 pairs of layers that could share parameters. With the option of having private spaces, this gives us $5^{19}=$19,073,486,328,125 possible MTL architectures. If we additionally consider soft sharing of parameters, the number of possible architectures grows infinite. It is obviously not feasible to search this space. 
Neural architecture search (NAS) \cite{Zoph:Le:17} typically requires learning from a large pool of experiments with different architectures. Searching for multi-task architectures via reinforcement learning \cite{Wong2018} or evolutionary approaches \cite{Liang:ea:18} can therefore be quite expensive. In this paper, we {\em jointly} learn a \emph{latent} multi-task architecture and task-specific models, paying a minimal computational cost over single task learning and standard multi-task learning (5-7\%~training time). We refer to this problem as {\em multi-task architecture learning}. In contrast to architecture \emph{search}, the overall meta-architecture is fixed and the model learns the optimal latent connections and pathways for each task. 
Recently, a few authors have considered multi-task architecture learning \cite{Misra2016,Meyerson2018}, but these papers only address a subspace of the possible architectures typically considered in neural multi-task learning, while other approaches at most consider a couple of architectures for sharing \cite{Soegaard:Goldberg:16,Peng2016b,Alonso:Plank:17}. In contrast, we introduce a framework that unifies previous approaches by introducing trainable parameters for all the components that differentiate multi-task learning approaches along the above dimensions.

\paragraph{Contributions} We present a novel meta-architecture (shown in Figure \ref{fig:architecture}) that generalizes several previous multi-task architectures, with an application to sequence tagging problems. Our meta-architecture enables multi-task architecture learning, i.e., learning (a) what layers to share between deep recurrent neural networks, but also (b) which parts of those layers to share, and with what strength, as well as (c) a mixture model of skip connections at the architecture's outer layer. We show that the architecture is a generalization of various multi-task \cite{Caruana:97,Soegaard:Goldberg:16,Misra2016} and transfer learning algorithms \cite{DaumeIII2007a}. We evaluate it on four tasks and across seven domains on OntoNotes 5.0 \cite{Weischedel2013}, where it consistently outperforms previous work on multi-task architecture learning, as well as common MTL approaches. Moreover, we study the task properties that predict gains and those that correlate with learning certain types of sharing.

\section{Multi-task Architecture Learning}

\begin{figure}[t!]
\centering
\includegraphics[width=\linewidth]{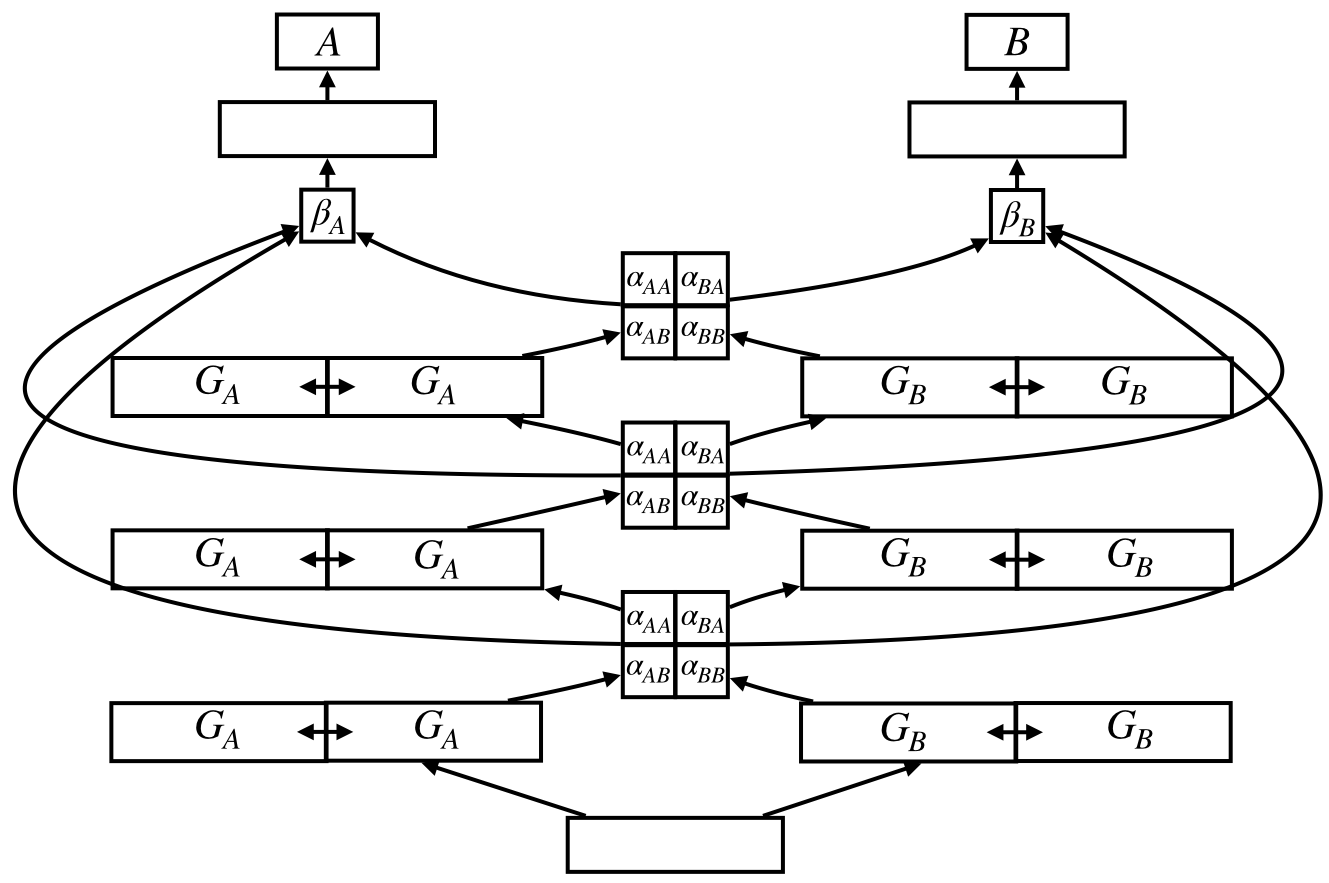}
\caption{A sluice meta-network with one main task $A$ and one auxiliary task $B$. It consists of a shared input layer (bottom), two task-specific output layers (top), and three hidden layers per task, each partitioned into two subspaces $G$ that are enforced to be orthogonal. $\alpha$ parameters control which subspaces are shared between main and auxiliary task, while $\beta$ parameters control which layer outputs are used for prediction. For simplicity, we do not index layers and subspaces. With two subspaces, each block $\alpha_{AA}, \alpha_{BA}, \ldots \in \mathbb{R}^{2\times 2}$. With three layers, $\beta_A, \beta_B \in \mathbb{R}^3$.}
\label{fig:architecture}
\end{figure}

We introduce a meta-architecture for multi-task architecture learning, which we refer to as a sluice network, sketched in Figure~\ref{fig:architecture} for the case of two tasks. The network learns to share parameters between $M$ neural networks---in our case, two deep recurrent neural networks (RNNs) \cite{hochreiter1997long}. The network can be seen as an end-to-end differentiable union of a set of sharing architectures with parameters controlling the sharing. By learning the weights of those sharing parameters (\emph{sluices}) jointly with the rest of the model, we arrive at a task-specific MTL architecture over the course of training.

The two networks $A$ and $B$ share an embedding layer associating the elements of an input sequence, in our case English words, with vector representations via word and character embeddings. The two sequences of vectors are then passed on to their respective inner recurrent layers. Each layer is divided into subspaces (by splitting the matrices in half), e.g., for network $A$ into $G_{A,1}$ and $G_{A,2}$, which allow the sluice network to learn task-specific and shared representations, if beneficial. The subspaces have different weights.

The output of the inner layer of network $A$ is then passed to its second layer, as well as to the second layer of network $B$. This traffic of information is mediated by a set of $\alpha$ and $\beta$ parameters similar to the way a sluice controls the flow of water. Specifically, the second layer of each network receives a combination of the output of the two inner layers weighted by the $\alpha$ parameters. Importantly, these $\alpha$ parameters are trainable and allow the model to learn whether to share or to focus on task-specific features in a subspace. Finally, a weighted combination of the outputs of the outer recurrent layers $G_{\cdot,3,\cdot}$ as well as the weighted outputs of the inner layers are mediated through $\beta$ parameters, which reflect a mixture over the representations at various depths of the multi-task architecture. 
In sum, {sluice networks} have the capacity to learn what layers and subspaces should be shared, and how much, as well as at what layers the meta-network has learned the best representations of the input sequences.

\paragraph{Matrix Regularization}  We cast learning what to share as a {\em matrix regularization} problem, following \cite{Jacobs:ea:08,Yang2017a}. Assume $M$ different tasks that are loosely related, with $M$ potentially non-overlapping datasets $\mathcal{D}_1,\ldots,\mathcal{D}_M$. Each task is associated with a deep neural network with $K$ layers $L_1, \ldots L_K$. For simplicity, we assume that all the deep networks have the same hyper-parameters at the outset. 
Let $W \in \mathbb{R}^{M \times D}$ be a matrix in which each row $i$ corresponds to a model $\theta_i$ with $D$ parameters. The loss that {sluice networks} minimize, with a penalty term $\Omega$, is then as follows: 
$\lambda_1 \mathcal{L}_1(\mathbf{f}(x;\theta_1),y_1)+\ldots+\lambda_M\mathcal{L}_M(\mathbf{f}(x;\theta_M),y_M)+\Omega$. 
 The loss functions $\mathcal{L}_i$ are cross-entropy functions of the form $-\sum_y p(y)\log q(y)$ where $y_i$ are the labels of task $i$. Note that sluice networks are not restricted to tasks with the same loss functions, but could also be applied to jointly learn regression and classification tasks. The weights $\lambda_i$ determine the importance of the different tasks during training. 
 We \emph{explicitly} add inductive bias to the model via the regularizer $\Omega$ below, but our model also \emph{implicitly} learns regularization through multi-task learning \cite{Caruana:93} mediated by the $\alpha$ parameters, while the $\beta$ parameters are used to learn the mixture functions $\mathbf{f}(\cdot)$, as detailed in the following.

\paragraph{Learning Matrix Regularizers} We now explain how updating $\alpha$ parameters can lead to different matrix regularizers. 
Each matrix $W$ consists of $M$ rows where $M$ is the number of tasks. Each row is of length $D$ with $D$ the number of model parameters. Subvectors $L_{m,k}$ correspond to the parameters of network $m$ at layer $k$. Each layer consists of two subspaces with parameters $G_{m,k,1}$ and $G_{m,k,2}$. Our meta-architecture is partly motivated by the observation that for loosely related tasks, it is often beneficial if only certain features in specific layers are shared, while many of the layers and subspaces remain more task-specific \cite{Soegaard:Goldberg:16}. We want to learn what to share while inducing models for the different tasks. For simplicity, we ignore subspaces at first and assume only two tasks $A$ and $B$. The outputs $h_{A, k, t}$ and $h_{B, k, t}$ of the $k$-th layer for time step $t$ for task $A$ and $B$ respectively interact through the $\alpha$ parameters (see Figure \ref{fig:architecture}). Omitting $t$ for simplicity, the output of the $\alpha$ layers is: 


\begin{equation}
\begin{bmatrix}
\widetilde{h}_{A, k}\\
\widetilde{h}_{B, k}
\end{bmatrix} = 
\begin{bmatrix}
           \alpha_{AA} & \alpha_{AB} \\
           \alpha_{BA} & \alpha_{BB}
         \end{bmatrix} 
\begin{bmatrix}
{h_{A, k}}^\top \: , & {h_{B, k}}^\top
\end{bmatrix}
\end{equation}

where $\widetilde{h}_{A, k}$ is a linear combination of the outputs that is fed to the $k+1$-th layer of task $A$, and
$\begin{bmatrix}a^\top , b^\top\end{bmatrix}$ designates the stacking of two vectors $a, b \in \mathbb{R}^D$ to a matrix $M \in \mathbb{R}^{2 \times D}$. Subspaces \cite{Virtanen2011,Bousmalis2016} should allow the model to focus on task-specific and shared features in different parts of its parameter space. Extending the $\alpha$-layers to include subspaces, for 2 tasks and 2 subspaces, we obtain an $\alpha$ matrix $\in \mathbb{R}^{4 \times 4}$ that not only controls the interaction between the layers of both tasks, but also between their subspaces:

\begin{equation}
\begin{split}
\begin{bmatrix}
\widetilde{h}_{A_1, k}\\
\vdots \\
\widetilde{h}_{B_2, k}
\end{bmatrix} = &
\begin{bmatrix}
           \alpha_{A_1A_1} & \ldots & \alpha_{B_2A_1} \\
           \vdots & \ddots & \vdots \\
           \alpha_{A_1B_2} & \ldots & \alpha_{B_2B_2}
\end{bmatrix} \\
& \begin{bmatrix}
{h_{A_1, k}}^\top \: , & \ldots ,&  {h_{B_2, k}}^\top
\end{bmatrix}
\end{split}
\end{equation}

where $h_{A_1, k}$ is the output of the first subspace of the $k$-th layer of task $A$ and $\widetilde{h}_{A_1, k}$ is the linear combination for the first subspace of task $A$.  The input to the $k+1$-th layer of task $A$ is then the concatenation of both subspace outputs: $h_{A, k} = \begin{bmatrix} \widetilde{h}_{A_1, k} \:, & \widetilde{h}_{A_2, k} 
\end{bmatrix}$. Different $\alpha$ weights correspond to different matrix regularizers $\Omega$, including several ones that have been proposed previously for multi-task learning. We review those in Section 3. For now just observe that if all $\alpha$-values are set to 0.25 (or any other constant), we obtain hard parameter sharing \cite{Caruana:93}, which is equivalent to a heavy $L_0$ matrix regularizer. 

\paragraph{Adding Inductive Bias}  Naturally, we can also add explicit  inductive bias to sluice networks by partially constraining the regularizer or adding to the learned penalty. Inspired by work on  shared-space component analysis \cite{Salzmann2010}, we add a penalty to enforce a division of labor and discourage redundancy between shared and task-specific subspaces. While the networks can theoretically learn such a separation, an explicit constraint empirically leads to better results and enables the sluice networks to take better advantage of subspace-specific $\alpha$-values. We introduce an orthogonality constraint \cite{Bousmalis2016} between the layer-wise subspaces of each model: $\Omega = \sum_{m=1}^M \sum_{k=1}^K \| {G_{m, k, 1}}^\top G_{m, k, 2} \|^2_F$, where $M$ is the number of tasks, $K$ is the number of layers, $\| \cdot \|^2_F$ is the squared Frobenius norm, and $G_{m, k, 1}$ and $G_{m, k, 2}$ are the first and second subspace respectively in the $k$-th layer of the $m$-th task model.

\paragraph{Learning Mixtures}

Many tasks form an implicit hierarchy of low-level to more complex tasks, with intuitive synergies between the low-level tasks and {\em parts of} the complex tasks. Rather than hard-coding this structure \cite{Soegaard:Goldberg:16,Hashimoto2016a}, we enable our model to learn hierarchical relations by associating different tasks with different layers if this is beneficial for learning. Inspired by advances in residual learning \cite{he2016deep}, we employ skip-connections from each layer, controlled using $\beta$ parameters. This layer acts as a mixture model, returning a mixture of expert predictions:

\begin{equation}
{\widetilde{h}_A}^\top = 
\begin{bmatrix}
           \beta_{A, 1} \\
           \cdots \\
           \beta_{A, k}
         \end{bmatrix}^\top
\begin{bmatrix}
{h_{A, 1}}^\top \: , & \ldots & {h_{A, k}}^\top
\end{bmatrix}
\end{equation}

where $h_{A, k}$ is the output of layer $k$ of model $A$, while $\widetilde{h}_{A, t}$ is the linear combination of all layer outputs of model $A$ that is fed into the final softmax layer.

\paragraph{Complexity} Our model only adds a minimal number of additional parameters compared to single-task models of the same architecture. In our experiments, we add $\alpha$ parameters between all task networks. As such, they scale linearly with the number of layers and quadratically with the number of tasks and subspaces, while $\beta$ parameters scale linearly with the number of tasks and the number of layers. For a sluice network with $M$ tasks, $K$ layers per task, and 2 subspaces per layer, we thus obtain $4KM^2$ additional $\alpha$ parameters and $KM$ $\beta$ parameters. Training sluice networks is not much slower than training hard parameter sharing networks, with only a 5--7\%~increase in training time.

\section{Prior Work as Instances of Sluice Networks} \label{sec:instantiations}

Our meta-architecture is very flexible and can be seen as a generalization over several existing algorithms for transfer and multi-task learning, including \cite{Caruana:97,DaumeIII2007a,Soegaard:Goldberg:16,Misra2016}. We show how to derive each of these below.

\begin{itemize}
\item {\bf Hard parameter sharing} between the two networks appears if all $\alpha$ values are set to the same constant \cite{Caruana:97,Collobert2008}.  
This is equivalent to a mean-constrained $\ell_0$-regularizer $\Omega(\cdot)=|\cdot|^{\bar{w_i}}_0$ and $\sum_i\lambda_i\mathcal{L}_i<1$. Since the penalty for not sharing a parameter is 1, it holds that if the sum of weighted losses is smaller than 1, the loss with penalty is always the highest when all parameters are shared.
\item  The $\ell_1/\ell_2$ {\bf group lasso} regularizer is $\sum^G_{g=1}||G_{1,i,g}||_2$, a weighted sum over the $\ell_2$ norms of the groups, often used to enforce subspace sharing \cite{Zhou:ea:10,Lozano:ea:12}. Our architecture learns a $\ell_1/\ell_2$ group lasso over the two subspaces (with the same degrees of freedom), when all $\alpha_{AB}$ and $\alpha_{BA}$-values are set to 0. When the outer layer $\alpha$-values are not shared, we get block communication between the networks.
\item The approach to domain adaptation in \cite{DaumeIII2007a}, commonly referred to as {\bf frustratingly easy domain adaptation}, which relies on a shared and a private space for each task or domain, can be encoded in sluice networks by setting all $\alpha_{AB}$- and $\alpha_{BA}$-weights associated with $G_{i,k,1}$ to 0, while setting all $\alpha_{AB}$-weights associated with $G_{i,k,2}$ to $\alpha_{BB}$, and $\alpha_{BA}$-weights associated with $G_{i,k,2}$ to $\alpha_{AA}$. Note that \citeauthor{DaumeIII2007a} \shortcite{DaumeIII2007a} discusses three subspaces. We obtain this space if we only share one half of the second subspaces across the two networks.
\item \citeauthor{Soegaard:Goldberg:16} \shortcite{Soegaard:Goldberg:16} propose a {\bf low supervision} model where only the inner layers of two deep recurrent works are shared. This is obtained using heavy mean-constrained $L_0$ regularization over the first layer $L_{i,1}$, e.g., $\Omega(W)=\sum_i^K||L_{i,1}||_0\mbox{ with }\sum_i\lambda_i\mathcal{L}(i)<1$, while for the auxiliary task, only the first layer $\beta$ parameter is set to 1.
\item \citeauthor{Misra2016} \shortcite{Misra2016} introduce {\bf cross-stitch networks} that have $\alpha$ values control the flow between layers of two convolutional neural networks. Their model corresponds to setting the $\alpha$-values associated with $G_{i,j,1}$ to be identical to those for $G_{i,j,2}$, and by letting all but the $\beta$-value associated with the outer layer be 0.
\end{itemize}

In our experiments, we include hard parameter sharing, low supervision, and cross-stitch networks as baselines. We do not report results for group lasso and frustratingly easy domain adaptation, which were consistently inferior, by some margin, on development data.\footnote{Note that frustratingly easy domain adaptation was not designed for MTL.}


\begin{figure}
\centering
\includegraphics[height=1.7in]{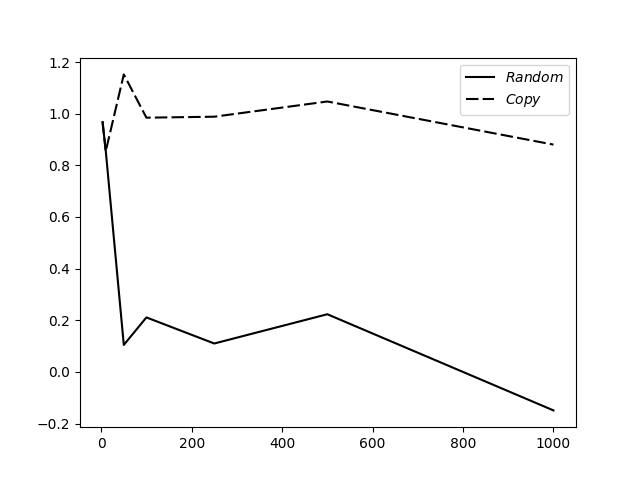}
\caption{\label{synth}The relative importance of the auxiliary task ($\frac{\alpha_{\mbox{\tiny BA}}}{\alpha_{\mbox{\tiny AA}}}$) over number of training instances. With more data, the network  learns {\em not}~to share, when auxiliary task is randomly relabeled ({\em Random}).}
\end{figure}

\section{Experiments}

\paragraph{A synthetic experiment} Our first experiment serves as a sanity check that our meta-architecture learns reasonable sharing architectures by learning $\alpha$ weights. We also want the $\alpha$ to adjust quickly in order not to slow down learning. We contrast two partially synthetic pairs of target and auxiliary data. In both cases, our target dataset is $n$ instances (sentences) from our part-of-speech tagging dataset (see details below). In the first scenario ({\em Random}), the auxiliary dataset is a random relabeling of the same $n$ instances. In the second scenario ({\em Copy}), the auxiliary dataset is a copy of the $n$ instances.

For {\em Random}, we would like our $\alpha$ parameters to quickly learn that the auxiliary task at best contributes with noise injection. Initializing our $\alpha$ parameters to equal weights (0.25), we therefore hope to see a quick drop in the relative importance of the auxiliary task, given by $\frac{\alpha_{\mbox{\tiny BA}}}{\alpha_{\mbox{\tiny AA}}}$. Seeing $n$ training instances, we expect this number to quickly drop, then stabilize to a slow decrease due to the reduced need for regularization with larger sample sizes.\footnote{The quick drop is the meta-architecture learning that the auxiliary data is much less useful than the target data; the slight decrease after the first drop is the reduced need for regularization due to lower variance with more data.}

For {\em Copy}, in contrast, we expect no significant change in the relative importance of the auxiliary task over $n$ training instances. We use the same hyperparameters as in our subsequent experiments (see below). The parameter settings are thus realistic, and not toy settings.

See Figure~\ref{synth} for the results of our experiment. The two curves show the expected contrast between an auxiliary task with an all-noise signal ({\em Random}) and an auxiliary task with a perfect, albeit redundant, signal ({\em Copy}). This experiment shows that our meta-architecture quickly learns a good sharing architecture in clear cases such as {\em Random} and {\em Copy}. We now proceed to test whether multi-task architecture learning also leads to empirical gains over existing approaches to multi-task learning. 


\paragraph{Data} As testbed for our experiments, we choose the OntoNotes 5.0 dataset \cite{Weischedel2013}, not only due to its high inter-annotator agreement \cite{Hovy2006}, but also because it enables us to analyze the generalization ability of our models across different tasks and domains. The OntoNotes dataset provides data annotated for an array of tasks across different languages and domains. We present experiments with the English portions of datasets, for which we show statistics in Table \ref{tab:ontonotes-5-stats}.

\begin{table}
\resizebox{\linewidth}{!}{%
\begin{tabular}{l c c c c c c c}
\toprule
& \multicolumn{7}{c}{\textbf{Domains}} \\\hline
 & \multicolumn{1}{c}{\textbf{bc}} & \multicolumn{1}{c}{\textbf{bn}} & \multicolumn{1}{c}{\multirow{1}{*}{\textbf{mz}}} & \multicolumn{1}{c}{\multirow{1}{*}{\textbf{nw}}} & \multicolumn{1}{c}{\multirow{1}{*}{\textbf{pc}}} & \multicolumn{1}{c}{\textbf{tc}} & \multicolumn{1}{c}{\multirow{1}{*}{\textbf{wb}}} \\
Train &  173289 &  206902  & 164217 &  878223  &297049  & 90403 & 388851 \\
Dev & 29957 &  25271 &  15421 & 147955  & 25206  & 11200  & 49393\\
Test &  35947 &  26424  & 17874 &  60756 &  25883 &  10916 &  52225\\
\bottomrule
\end{tabular}
}
\caption{Number of tokens for each domain in the OntoNotes 5.0 dataset. }
\label{tab:ontonotes-5-stats}
\end{table}

\begin{table}
\resizebox{\linewidth}{!}{%
\begin{tabular}{l | l l l l l}
\toprule
\sc{Words} & Abramov & had & a & car & accident \\
\midrule
\sc{CHUNK} & O & B-VP & B-NP & I-NP & I-NP \\ 
\sc{NER} & B-PERSON & O & O & O & O \\
\sc{SRL} & B-ARG0 & B-V & B-ARG1 & I-ARG1 & I-ARG1 \\
\sc{POS} & NNP & VBD & DT & NN & NN\\
\bottomrule
\end{tabular}
}
\caption{Example annotations for {\sc CHUNK}, {\sc NER}, {\sc SRL}, and {\sc POS}.}
\label{tab:task-annotations}
\end{table}

\paragraph{Tasks} In MTL, one task is usually considered the main task, while other tasks are used as auxiliary tasks to improve performance on the main task. As main tasks, we use chunking ({\sc CHUNK}), named entity recognition ({\sc NER}), and a simplified version of semantic role labeling ({\sc SRL}) where we only identify headwords\footnote{We do this to keep pre-processing for SRL minimal.}
, and pair them with part-of-speech tagging ({\sc POS}) as an auxiliary task, following \cite{Soegaard:Goldberg:16}. Example annotations for each task can be found in Table \ref{tab:task-annotations}.

\begin{table*}[!htb]
\centering
\begin{tabular}{c l c c c c c c c c}
\toprule
\multicolumn{10}{c}{In-domain results}\\\midrule
& System & bc & bn & mz & nw & pt & tc & wb & Avg \\
\midrule
\multirow{4}{*}{\rotatebox[origin=c]{90}{Baselines}} & Single task & 90.80 & 92.20 & 91.97 & 92.76 & 97.13 & 89.84 & 92.95 & 92.52 \\
& Hard sharing & 90.31 & 91.73 & 92.33 & 92.22 & 96.40 & 90.59 & 92.84 & 92.35 \\
& Low supervision & 90.95 & 91.70 & 92.37 & 93.40 & 96.87 & 90.93 & 93.82 & 92.86 \\
& Cross-stitch nets & 91.40 & 92.49 & 92.59 & 93.52 & 96.99 & \textbf{91.47} & 94.00 & 93.21 \\
\midrule
Ours & Sluice network & \textbf{91.72} & \textbf{92.90} & \textbf{92.90} & \textbf{94.25} & \textbf{97.17} & 90.99 & \textbf{94.40} & \textbf{93.48} \\\midrule
\multicolumn{10}{c}{Out-of-domain results}\\\midrule
\multirow{4}{*}{\rotatebox[origin=c]{90}{Baselines}} & Single task & 85.95 & 87.73 & 86.81 & 84.29 & 90.91 & 84.55 & 73.36 & 84.80 \\
& Hard sharing & 86.31 & 87.73 & 86.96 & 84.99 & 90.76 & 84.48 & 73.56 & 84.97 \\
& Low supervision & 86.53 & 88.39 & 87.15 & 85.02 & 90.19 & 84.48 & 73.24 & 85.00 \\
& Cross-stitch nets & 87.13 & 88.40 & 87.67 & 85.37 & 91.65 & \textbf{85.51} & 73.97 & 85.67 \\
\midrule
Ours & Sluice network & \textbf{87.95} & \textbf{88.95} & \textbf{88.22} & \textbf{86.23} & \textbf{91.87} & 85.32 & \textbf{74.48} & \textbf{86.15} \\\bottomrule \\
\end{tabular}
\caption{Accuracy scores for chunking on in-domain and out-of-domain test sets with POS as auxiliary task. Out-of-domain results for each target domain are averages across the 6 remaining source domains. Error reduction vs. single task: 12.8\%~(in-domain), 8.9\% (out-of-domain); vs. hard parameter sharing: 14.8\% (in-domain).}
\label{tab:results-chunking-pos}
\end{table*}

\paragraph{Model} We use a state-of-the-art BiLSTM-based sequence labeling model \cite{Plank2016a} as the building block of our model. The BiLSTM consists of 3 layers with 100 dimensions that uses 64-dimensional word and 100-dimensional character embeddings, which are both randomly initialized. The output layer is an MLP with a dimensionality of 100. We initialize $\alpha$ parameters with a bias towards one source subspace for each direction and initialize $\beta$ parameters with a bias towards the last layer\footnote{We experimented with different initializations for $\alpha$ and $\beta$ parameters and found these to work best.}. We have found it most effective to apply the orthogonality constraint only to the weights associated with the LSTM inputs.

\paragraph{Training and Evaluation} We train our models with stochastic gradient descent (SGD), an initial learning rate of 0.1, and learning rate decay\footnote{We use SGD as \citeauthor{Soegaard:Goldberg:16} \shortcite{Soegaard:Goldberg:16} also employed SGD. Adam yielded similar performance differences.}. During training, we uniformly sample from the data for each task. We perform early stopping with patience of 2 based on the main task and hyperparameter optimization on the in-domain development data of the newswire domain. We use the same hyperparameters for all comparison models across all domains. We train our models on each domain and evaluate them both on the in-domain test set (Table \ref{tab:results-chunking-pos}, top) as well as on the test sets of all other domains (Table \ref{tab:results-chunking-pos}, bottom) to evaluate their out-of-domain generalization ability. Note that due to this set-up, our results are not directly comparable to the results reported in \cite{Soegaard:Goldberg:16}, who only train on the WSJ domain and use OntoNotes 4.0.

\paragraph{Baseline Models} As baselines, we compare against i) a single-task model only trained on chunking; ii) the low supervision model \cite{Soegaard:Goldberg:16}, which predicts the auxiliary task at the first layer; iii) an MTL model based on hard parameter sharing \cite{Caruana:93}; and iv) cross-stitch networks \cite{Misra2016}. We compare these against our complete sluice network with subspace constraints and learned $\alpha$ and $\beta$ parameters. We implement all models in DyNet \cite{neubig2017dynet} and make our code available at \url{https://github.com/sebastianruder/sluice-networks}.

We first assess how well sluice networks perform on in-domain and out-of-domain data compared to the state-of-the-art. We evaluate all models on chunking with POS tagging as auxiliary task.

\paragraph{Chunking results} We show results on in-domain and out-of-domain tests sets in Table \ref{tab:results-chunking-pos}. On average, sluice networks significantly outperform all other model architectures on both in-domain and out-of-domain data and perform best for all domains, except for the telephone conversation (tc) domain, where they are outperformed by cross-stitch networks. The performance boost is particularly significant for the out-of-domain setting, where sluice networks add more than 1 point in accuracy compared to hard parameter sharing and almost .5 compared to the strongest baseline on average, demonstrating that sluice networks are particularly useful to help a model generalize better. In contrast to previous studies on MTL \cite{Alonso:Plank:17,Bingel:ea:17,Augenstein2018}, our model also consistently outperforms single-task learning. Overall, this demonstrates that our meta-architecture for learning which parts of multi-task models to share, with a small set of additional parameters to learn, can achieve significant and consistent improvements over strong baseline methods.

\paragraph{NER and SRL} We now evaluate sluice nets on NER with POS tagging as auxiliary task and simplified semantic role labeling with POS tagging as auxiliary task. We show results in Table \ref{tab:results-ner-srl}. Sluice networks outperform the comparison models for both tasks on in-domain test data and successfully generalize to out-of-domain test data on average. They yield the best performance on 5 out of 7 domains and 4 out of 7 domains for NER and semantic role labeling.

\begin{table*}[!htb]
\centering
\begin{tabular}{c l c c c c c c c c}
\toprule
\multicolumn{9}{c}{Named entity recognition}\\\midrule
& System & nw (ID) & bc & bn & mz & pt & tc & wb & OOD Avg\\\midrule
\multirow{4}{*}{\rotatebox[origin=c]{90}{Baselines}} & Single task & 95.04 & 93.42 & 93.81 & 93.25 & 94.29 & 94.27 & 92.52 & 93.59 \\
& Hard sharing & 94.16 & 91.36 & 93.18 & 93.37 & \textbf{95.17} & 93.23 & \textbf{92.99} & 93.22 \\
& Low supervision & 94.94 & 91.97 & 93.69 & 92.83 & 94.26 & 93.51 & 92.51 & 93.13 \\
& Cross-stitch nets & 95.09 & 92.39 & 93.79 & 93.05 & 94.14 & 93.60 & 92.59 & 93.26\\
\midrule
Ours & Sluice network & \textbf{95.52} & \textbf{93.50} & \textbf{94.16} & \textbf{93.49} & 93.61 & \textbf{94.33} & 92.48 & \textbf{93.60} \\
\midrule
\multicolumn{9}{c}{Simplified semantic role labeling}\\\midrule
\multirow{4}{*}{\rotatebox[origin=c]{90}{Baselines}} & Single task & 97.41 & \textbf{95.67} & 95.24 & 95.86 & 95.28 & 98.27 & 97.82 & 96.36 \\
& Hard sharing & 97.09 & 95.50 & 95.00 & 95.77 & \textbf{95.57} & 98.46 & 97.64 & 96.32\\
& Low supervision & 97.26 & 95.57 & 95.09 & 95.89 & 95.50 & 98.68 & 97.79 & 96.42 \\
& Cross-stitch nets & 97.32 & 95.44 & 95.14 & 95.82 & \textbf{95.57} & \textbf{98.69} & 97.67 & 96.39 \\
\midrule
Ours & Sluice network & \textbf{97.67} & 95.64 & \textbf{95.30} & \textbf{96.12} & 95.07 & 98.61 & \textbf{98.01} & \textbf{96.49}\\
\bottomrule \\
\end{tabular}
\caption{Test accuracy scores for different target domains with nw as source domain for named entity recognition (main task) and simplified semantic role labeling with POS tagging as auxiliary task for baselines and Sluice networks. ID: in-domain. OOD: out-of-domain.}
\label{tab:results-ner-srl}
\end{table*}

\begin{table}
\centering
\begin{tabular}{l c c c c}
\toprule
System & {\sc chunk} & {\sc ner} & {\sc srl} & {\sc pos} \\\midrule
Single task & \textbf{89.30} & 94.18 & 96.64 & 88.62 \\
Hard param. & 88.30 & 94.12 & \textbf{96.81} & 89.07\\
Low super. & 89.10 & 94.02 & 96.72 & 89.20\\
\midrule
Sluice net & 89.19 & \textbf{94.32} & 96.67 & \textbf{89.46}\\
\bottomrule
\end{tabular}
\caption{All-tasks experiment: Test accuracy scores for each task with nw as source domain averaged across all target domains.}
\label{tab:results-all-tasks}
\end{table}

\paragraph{Joint model} Most work on MTL for NLP uses a single auxiliary task \cite{Bingel:ea:17,Alonso:Plank:17}. 
In this experiment, we use one sluice network to jointly learn our four tasks on the newswire domain and show results in Table \ref{tab:results-all-tasks}.

Here, the low-level POS tagging and simplified SRL tasks are the only ones that benefit from hard parameter sharing highlighting that hard parameter sharing by itself is not sufficient for doing effective multi-task learning with semantic tasks. We rather require task-specific layers that can be used to transform the shared, low-level representation into a form that is able to capture more fine-grained task-specific knowledge. 
Sluice networks outperform single task models for all tasks, except chunking and achieve the best performance on 2/4 tasks in this challenging setting.

\begin{table*}
\centering
\resizebox{\textwidth}{!}{
\begin{tabular}{l l c c c c c c c c}
\toprule
\parbox{1.9cm}{Task sharing} & Layer sharing & bc & bn & mz & nw & pt & tc & wb & Avg\\
\midrule
\multirow{3}{*}{constant $\alpha$ (hard)} & Concatenation & 86.70 & 88.24 & 87.20 & 85.19 & 90.64 & 85.33 & 73.75 & 85.29 \\
& Skip-connections ($\beta = 1$) & 86.65 & 88.10 & 86.82 & 84.91 & 90.92 & 84.89 & 73.62 & 85.13\\
& Mixture (learned $\beta$) & 86.59 & 88.03 & 87.19 & 85.12 & 90.99 & 84.90 & 73.48 & 85.19\\
\midrule
\multirow{4}{*}{learned $\alpha$ (soft)} & Concatenation & 87.37 & 88.94 & 87.99 & 86.02 & \textbf{91.96} & \textbf{85.83} & 74.28 & 86.05\\
& Skip-connections & 87.08 & 88.62 & 87.74 & 85.77 & 91.92 & 85.81 & 74.04 & 85.85\\
& Mixture & 87.10 & 88.61 & 87.71 & 85.44 & 91.61 & 85.55 & 74.09 & 85.73\\
& Mixture + subspaces & \textbf{87.95} & \textbf{88.95} & \textbf{88.22} & \textbf{86.23} & 91.87 & 85.32 & \textbf{74.48} & \textbf{86.15}\\
\bottomrule \\
\end{tabular}
}
\caption{Ablation. Out-of-domain scores for Chunking with POS as auxiliary task
, averaged across the 6 source domains.}
\label{tab:ablation}
\end{table*}

\section{Analysis}\label{sec:ablation-analysis}

To better understand the properties and behavior of our meta-architecture, we conduct a series of analyses and ablations.

\paragraph{Task Properties and Performance} \citeauthor{Bingel:ea:17} \shortcite{Bingel:ea:17} correlate meta-characteristics of task pairs and gains compared to hard parameter sharing across a large set of NLP task pairs. Similarly, we correlate various meta-characteristics with error reductions and $\alpha,\beta$ values. Most importantly, we find that a) multi-task learning gains, also in sluice networks, are higher when there is less training data; and b) sluice networks learn to share more when there is more variance in the training data (cross-task $\alpha$s are higher, intra-task $\alpha$s lower). Generally, $\alpha$ values at the inner layers correlate more strongly with meta-characteristics than $\alpha$ values at the outer layers.

\paragraph{Ablation Analysis} Different types of sharing may be more important than others. In order to analyze this, we perform an ablation analysis in Table~\ref{tab:ablation}. We investigate the impact of i) the $\alpha$ parameters; ii) the $\beta$ parameters; and iii) the division into subspaces with an orthogonality penalty. We also evaluate whether concatenation of the outputs of each layer is a reasonable alternative to our mixture model. Overall, we find that learnable $\alpha$ parameters are preferable over constant $\alpha$ parameters. Learned $\beta$ parameters marginally outperform skip-connections in the hard parameter sharing setting, while skip-connections are competitive with learned $\beta$ values in the learned $\alpha$ setting.

In addition, modeling subspaces explicitly helps for almost all domains. To our knowledge, this is the first time that subspaces within individual LSTM layers have been shown to be beneficial.\footnote{\citeauthor{Liu2017} \shortcite{Liu2017} induce subspaces between separate LSTM layers.}. 
Being able to effectively partition LSTM weights opens the way to research in inducing more structured neural network representations that encode task-specific priors. Finally, concatenation of layer outputs is a viable form to share information across layers as has also been demonstrated by recent models such as DenseNet \cite{Huang2017}.

\paragraph{Analysis of $\alpha$ values} We analyze the final $\alpha$ weights in the sluice networks for Chunking, NER, and SRL, trained with newswire as training data. We find that a) for the low-level simplified SRL, there is more sharing at inner layers, which is in line with \citeauthor{Soegaard:Goldberg:16} \shortcite{Soegaard:Goldberg:16}, while Chunking and NER also rely on the outer layer, and b) more information is shared from the more complex target tasks than vice versa. 

\paragraph{Analysis of $\beta$ values} Inspecting the $\beta$ values for the all-tasks sluice net in Table \ref{tab:results-all-tasks}, we find that all tasks place little emphasis on the first layer, but prefer to aggregate their representations in different later layers of the model: The more semantic NER and chunking tasks use the second and third layer to a similar extent, while for POS tagging and simplified SRL the representation of one of the two later layers dominates the prediction.

\section{Related Work}

\emph{Hard parameter sharing} \cite{Caruana:93} is easy to implement, reduces overfitting, but is only guaranteed to work for (certain types of) closely related tasks~\cite{Baxter:00,Maurer:07}. \citeauthor{Peng2016b} \shortcite{Peng2016b} apply a variation of hard parameter sharing to multi-domain multi-task sequence tagging with a shared CRF layer and domain-specific projection layers. \citeauthor{Yang2016} \shortcite{Yang2016} use hard parameter sharing to jointly learn different sequence-tagging tasks across languages. \citeauthor{Alonso:Plank:17} \shortcite{Alonso:Plank:17} explore a similar set-up, but sharing is limited to the initial layer. In all three papers, the amount of sharing between the networks is fixed in advance. 

In \emph{soft parameter sharing} \cite{Duong:ea:15}, each task has separate parameters and separate hidden layers, as in our architecture, but the loss at the outer layer is regularized by the current distance between the models. 
\citeauthor{Kumar2012} \shortcite{Kumar2012} and 
\citeauthor{Maurer2013} \shortcite{Maurer2013} enable {\em selective sharing} by allowing task predictors to select from sparse parameter bases for homogeneous tasks. 
\citeauthor{Soegaard:Goldberg:16} \shortcite{Soegaard:Goldberg:16} 
show that low-level tasks, i.e. syntactic tasks typically used for preprocessing such as POS tagging and NER, should be supervised at lower layers when used as auxiliary tasks.

Another line of work looks into separating the learned space into a private (i.e. task-specific) and shared space~\cite{Salzmann2010,Virtanen2011} to more explicitly capture the difference between task-specific and cross-task features. Constraints are enforced to prevent the models from duplicating information. \citeauthor{Bousmalis2016} \shortcite{Bousmalis2016} use shared and private encoders regularized with orthogonality and similarity constraints for domain adaptation for computer vision. \citeauthor{Liu2017} \shortcite{Liu2017} use a similar technique for sentiment analysis. In contrast, we do not limit ourselves to a predefined way of sharing, but let the model learn which parts of the network to share using latent variables, the weights of which are learned in an end-to-end fashion. \citeauthor{Misra2016} \shortcite{Misra2016}, focusing on applications in computer vision, consider a small subset of the sharing architectures that are learnable in sluice networks, i.e., {\em split architectures}, in which two $n$-layer networks share the innermost $k$ layers with $0\leq k\leq n$, but learn $k$ with a mechanism  very similar to $\alpha$-values. 

Our method is also related to the classic mixture-of-experts layer \cite{Jacobs1991}. In contrast to this approach, our method is designed for multi-task learning and thus encourages a) the sharing of parameters between different task ``experts'' if this is beneficial as well as b) differentiating between low-level and high-level representations.

\section{Conclusion}

We introduced sluice networks, a meta-architecture for multi-task architecture search. In our experiments across four tasks and seven different domains, the meta-architecture consistently improved over strong single-task learning, architecture learning, and multi-task learning baselines. We also showed how our meta-architecture can learn previously proposed architectures for multi-task learning and domain adaptation. 

\section*{Acknowledgements}
Sebastian is supported by Irish Research Council Grant Number EBPPG/2014/30 and Science Foundation Ireland Grant Number SFI/12/RC/2289, co-funded by the European Regional Development Fund.

\bibliography{mtl_sharing}
\bibliographystyle{aaai}

\end{document}